%% file: main.tex
\documentclass[10pt,twocolumn,letterpaper]{article}

\usepackage[pagenumbers]{cvpr} 

\usepackage[accsupp]{axessibility}
\usepackage{graphicx}
\usepackage{amsmath}
\usepackage{amssymb}
\usepackage{amsthm}

\usepackage{booktabs}
\usepackage{multirow}
\usepackage{makecell}
\usepackage{rotating}

\usepackage{pifont}
\usepackage{bm}
\usepackage{txfonts}

\usepackage{algorithm}
\usepackage{algorithmicx}
\usepackage{algpseudocode}
\usepackage[dvipsnames]{xcolor}

\usepackage[pagebackref,breaklinks,colorlinks]{hyperref}

\newcommand{\name}{DLB}

\newcommand{\figref}[1]{Figure \ref{#1}}
\newcommand{\tabref}[1]{Table \ref{#1}}

\definecolor{moco}{RGB}{33, 130, 129}
\definecolor{myRed}{RGB}{195,10,10}
\definecolor{myGreen}{RGB}{55,149,73}

\newcommand{\cmark}{\textcolor{myRed}{\ding{51}}}
\newcommand{\xmark}{\textcolor{myGreen}{\ding{55}}}
\newcommand{\upmark}{\textcolor[RGB]{215,35,35}{$\uparrow$}}

\def\cvprPaperID{****} 
\def\confName{CVPR}
\def\confYear{2022}

\begin{document}

\title{Self-Distillation from the Last Mini-Batch for Consistency Regularization}

\author{
Yiqing Shen\textsuperscript{1*\dag}, 
Liwu Xu\textsuperscript{2*},  
Yuzhe Yang\textsuperscript{2}, 
Yaqian Li\textsuperscript{2\ddag}, 
Yandong Guo\textsuperscript{2}\\
\textsuperscript{1}Shanghai Jiao Tong University, \quad
\textsuperscript{2}OPPO Research Institute\\
{\tt\small 
shenyq@sjtu.edu.cn,
\{xuliwu, liyaqian\}@oppo.com},
{\tt\small ippllewis@gmail.com, yandong.guo@live.com}
}

\maketitle

\def\thefootnote{*}\footnotetext{Equal Contribution.~\textsuperscript{\ddag}Corresponding Author.}

\def\thefootnote{\dag}\footnotetext{This work was done during the internship at OPPO Research Institute.}

\input{1-abstract}
\input{2-intro}
\input{3-related}

\input{4-method}

\input{5-exp}

\input{6-conclusion}

{\small
\bibliographystyle{ieee_fullname}
\bibliography{7-ref}
}

\end{document}

%% file: 1-abstract.tex
\begin{abstract}
Knowledge distillation (KD) shows a bright promise as a powerful regularization strategy to boost generalization ability by leveraging learned sample-level soft targets. Yet, employing a complex pre-trained teacher network or an ensemble of peer students in existing KD is both time-consuming and computationally costly. Various self KD methods have been proposed to achieve higher distillation efficiency. However, they either require extra network architecture modification or are difficult to parallelize. To cope with these challenges, we propose an efficient and reliable self-distillation framework, named Self-\underline{D}istillation from \underline{L}ast Mini-\underline{B}atch (\name). Specifically, we rearrange the sequential sampling by constraining half of each mini-batch coinciding with the previous iteration. Meanwhile, the rest half will coincide with the upcoming iteration. Afterwards, the former half mini-batch distills on-the-fly soft targets generated in the previous iteration. Our proposed mechanism guides the training stability and consistency, resulting in robustness to label noise. Moreover, our method is easy to implement, without taking up extra run-time memory or requiring model structure modification. Experimental results on three classification benchmarks illustrate that our approach can consistently outperform state-of-the-art self-distillation approaches with different network architectures. Additionally, our method shows strong compatibility with augmentation strategies by gaining additional performance improvement. The code is available at \url{https://github.com/Meta-knowledge-Lab/DLB}.
\end{abstract}

%% file: 2-intro.tex
\section{Introduction}

\begin{table*}[!ht]
\caption{
A comparison with the state-of-the-arts in terms of computational cost and smoothed granularity. We compare our method with label smoothing regularization \cite{LSR}, teacher-free knowledge distillation (Tf-KD$_{self}$, Tf-KD$_{reg}$) \cite{tf-kd}, class-wise self-knowledge distillation (CS-KD) \cite{cs-kd}, progressive self-knowledge distillation (PS-KD) \cite{ps-kd}, memory-replay knowledge distillation (Mr-KD) \cite{Mr-KD}, data-distortion guided self-knowledge distillation (DDGSD) \cite{r-aug}, be your own teacher (BYOT) \cite{be-your-own-teacher}.
}\label{table:method_comparison}
\centering
\resizebox{\linewidth}{!}{
\begin{tabular}{l|ccccccccc} 
\toprule
Characteristic& LSR & Tf-KD$_{self}$ & Tf-KD$_{reg}$ & CS-KD & PS-KD & Mr-KD & DDGSD & BYOT & \textbf{Ours}\\
\hline
Sample-level smoothing& \xmark  & \cmark & \cmark & \cmark & \cmark & \cmark & \cmark & \cmark & \cmark\\
No pre-trained teacher& \cmark & \xmark & \cmark & \cmark & \cmark & \cmark & \cmark & \cmark & \cmark\\
No Architecture modification& \cmark & \cmark & \cmark & \cmark & \cmark & \cmark & \cmark & \xmark & \cmark \\
Forward times per batch & 1 & 2 & 1 & 2 & 2 & $\ge2$ & 2 & 1 & 1 \\
Backward times per batch & 1 & 1 & 1 & 2 & 1 & 1& 2 & $\ge$2 & 1 \\
Number of involved networks & 1 & 2 & 1 & 1 & 2 & 1 & 1 & 1 & 1 \\
Label update frequency & - & epoch & - & - & epoch & epoch & - & - & batch \\
\bottomrule
\end{tabular}
}
\end{table*}

Knowledge Distillation (KD), first introduced by Bucilua et al. \cite{kd1}, was later popularized by Hinton et al. \cite{kd2}. Numerous previous researches have demonstrated the success of KD in various learning tasks to boost the generalization ability. For example, in the case of network compression, the two-stage offline KD is widely used to transfer dark knowledge from a cumbersome pre-trained model to a light student model that learns from teachers' intermediate feature maps \cite{fitnet}, logits \cite{kd2}, attention maps \cite{kd-attention}, or auxiliary outputs \cite{be-your-own-teacher}. However, training a high-capacity teacher network heavily relies on large computational sources and run-in memory. To alleviate the time-consuming preparation of static teachers, online distillation is introduced \cite{dml}, where an ensemble of peer students learns mutually from each other. Online KD achieves equivalent performance improvement, compared with offline KD, with higher computational efficiency. Thus, this line is subsequently extended by many following works to a more capable self-ensemble teacher \cite{one,okddip,kdcl,pcl}. Other applications of KD include semi-supervised learning, domain adaptation, transfer learning and etc \cite{kd_app_3,kd_app_1,kd_app_2}. The main scope of this paper focuses on the KD paradigm itself.

Conventional KD approaches, both online and offline, have achieved satisfying empirical performance \cite{performance_improve}. Yet, existing KD approaches suffer from an obstacle in the low knowledge transferring efficiency \cite{r-aug}. Additionally, high computation and run-in memory costs restrict their deployment onto the end devices, such as mobile phones, digital cameras \cite{Moonshine}. To cope with these limitations, self-knowledge distillation has received increasing popularity, which enables a student model to distill knowledge from itself. The absence of a complex pre-trained teacher and an ensemble of peer students in self-KD contributes to a marginal improvement in the training efficiency. 

One popular formulation of self KD, such as Be Your Own Teacher (BYOT), requires heavy network architecture modifications, which largely increases their difficulty to generalize onto various network structures \cite{be-your-own-teacher,byot2,byot3}. In another line, history information, including previous training logits or model snapshots, is utilized to construct a virtual teacher for extra supervision signals as self distillation. Initially, born again networks (BAN) sequentially distill networks with identical parameters as its last generation \cite{ban}. An advancement achieved by snapshot distillation is to take the secondary information from the prior mini-generation i.e. a couple of epochs, within one generation \cite{snapshot}. This virtual teacher update frequency is further improved to epoch-level in progressive self-knowledge distillation \cite{ps-kd} and learning with retrospection \cite{lwr}. Yet, existing self KD methods have the following setbacks to be tackled. Firstly, the most instant information from the last iteration is discarded. Moreover, storing a snapshot of past models consumes an extra run-in memory cost, and subsequently increases the difficulty to parallelize \cite{snapshot}. Finally, computation of the gradient in each time backward prorogation is associated with twice the forward process on each batch of data, resulting in a computational redundancy and low computational efficiency.

To address these challenges in existing self KD methods, we propose a simple but efficient self distillation approach, named as Self-\underline{D}istillation from \underline{L}ast Mini-\underline{B}atch (\name). Compared with existing self KD approaches, DLB is computationally efficient and saves the run-in memory by only storing the soft targets generated in the last mini-batch backup, resulting in its simplicity in deployment and parallelization. Every forward process of data instances is associated with a once backpropagation process, mitigating the computational redundancy. Major differences compared with the state-of-the-arts are summarized in \tabref{table:method_comparison}. DLB produces on-the-fly sample-level smoothed labels for self-distillation. Leveraging the soft predictions from the last iteration, our method provides the most instant distillation for each training sample. The success of DLB is attributed to the distillation from the most immediate historically generated soft targets to impose training consistency and stability. To be more specific, the target network plays a dual role as teacher and student in each mini-batch during the training stage. As a teacher, it provides soft targets to regularize itself in the next iteration. As a student, it distills smoothed labels generated from the last iteration and minimizes the supervision learning objective e.g. the cross-entropy loss.

We empirically illustrate the comprehensive effectiveness of our methods on three benchmark datasets, namely CIFAR-10 \cite{cifar}, CIFAR-100 \cite{cifar}, TinyImageNet. Our approach is both task-agnostic and model-agnostic i.e. with no requirement of model topological modifications. We select six representative backbone CNNs for evaluations, including ResNet-18, ResNet-110 \cite{resnet}, VGG-16, VGG-19 \cite{vgg}, DenseNet \cite{densenet}, WideResNet \cite{wrn}. Experimental results demonstrate that our \name~ can consistently improve the generalization ability. We also test the robustness of DLB on corrupted data. Training consistency and stability imposed by DLB on corrupted data results in higher generalization ability. 

\begin{figure*}[!htb]
\centering
\includegraphics[width=1.0\linewidth]{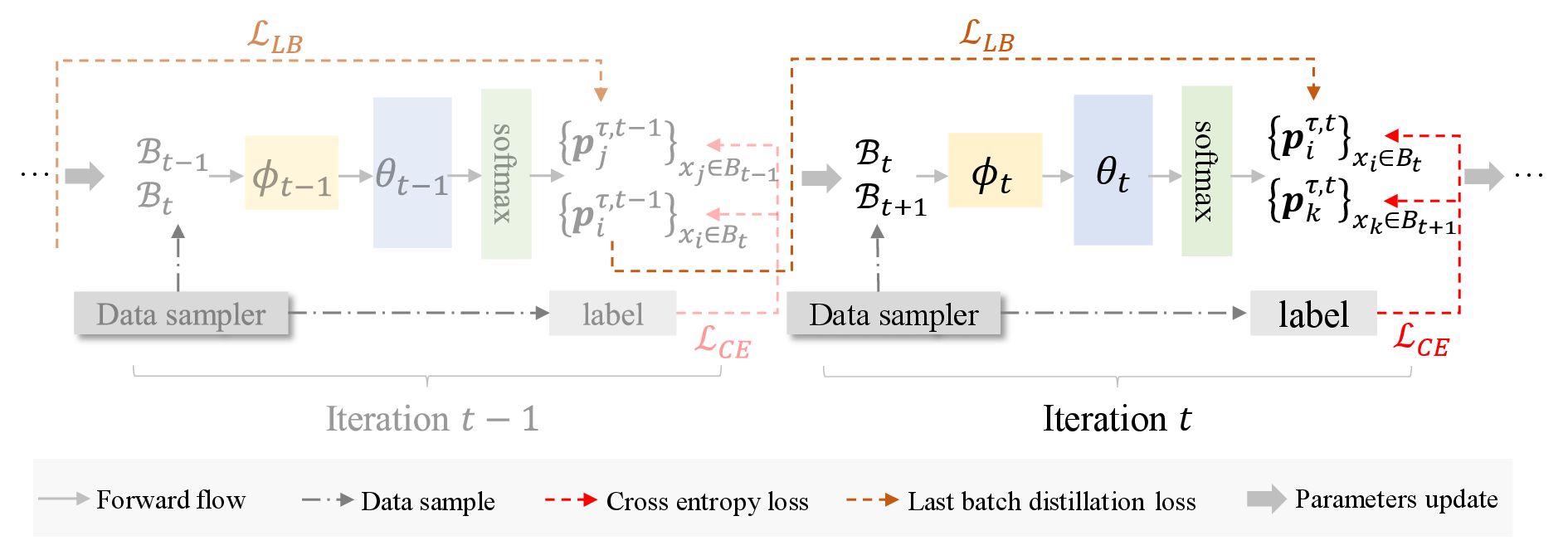}
\caption{The overall architecture of our \name. We write $\mathcal{B}_t$, $\phi_t$, $\theta_t$ for a mini-batch of data samples, random augmentation and the trainable parameters indexed in the $t$\textsuperscript{th} iteration.}
\label{fig:framework}
\end{figure*}

The major contributions are three-fold.
\begin{itemize}
    \item We propose a simple but efficient consistency regularization scheme based on self-knowledge distillation, named as \name. With no network architecture modifications, our method requires very few additional computation costs as well as the run-time memory to implement. Utilizing the latest update from the last iteration, our DLB is easy to implement for parallelization. Notably, the proposed method is also both model-agnostic and task-agnostic.
    \item  Comprehensive experimental results on three popular classification benchmarks illustrate consistent generalization improvements on different models. We also empirically demonstrate the compatibility of DLB with various augmentation strategies. 

    \item We systemically analyze the impact of our method on the training dynamics. Concretely, the success of its regularization effect is attributed to guidance towards training consistency, by leveraging on-the-fly sample-level smooth labels. The consistency effect is further amplified under label corruption settings, showing strong robustness to label noise. These empirical finds may shed light on a new direction to understand the effect of knowledge distillation.
\end{itemize}

%% file: 3-related.tex
\section{Related Works}

\paragraph{Knowledge distillations.} Knowledge distillation (KD) targets to transfer `knowledge', such as logits, or intermediate feature maps, from a high-capability teacher model to a lightweight student network \cite{kd1,kd2}. Despite its competitive performance improvement to generalization, pre-training a complex teacher model requires extra training time and computational cost. Another way to form an economic distillation is called mutual learning, also known as online distillation, where an ensemble of students learn mutually from each other \cite{dml}. This idea was popularized by many following works \cite{one,okddip,pcl}. But, the optimization in peer learning involves multiple numbers of networks, which requires extra memory to store all parameters. 

\paragraph{Self Knowledge Distillation.} To enhance efficiency and effectiveness in knowledge transferring, self knowledge distillation (SKD) is proposed to utilize knowledge from itself, without the involvement of an extra networks \cite{open_challenge}. There are three popular ways to construct a SKD model i.e., 1) data distortion based self-distillation \cite{r-aug,r-augv2}, 2) use history information as a virtual teacher, 3) distilling across auxiliary heads \cite{be-your-own-teacher,msd}. However, the first one marginally relies on the augmentation efficiency. The second one misses the latest update from the last mini-batch. And the last kind requires heavy network architecture modifications, which increase its difficulty for deployment.

\paragraph{Distillations as regularization.} KD is extensively used in many tasks, such as model compression, semi-supervised learning, domain adaptation and etc \cite{kd_app_1,kd_app_2,kd_app_3}. However, theoretical analysis of the success of KD remains a big challenge. Recently, Yuan et al. attributed the success of KD to its regularization effect as providing sample-level soft targets from the LSR perspective \cite{tf-kd}. It reveals the large promise of applying KD to the regularization domain. In this line, class-wise self-knowledge distillation (CS-KD) is designed by evacuating a consistency between predictions of two batches of samples identified with the same category \cite{cs-kd}. Progressive self-knowledge distillation (PS-KD), more similar to our work, progressively distills past knowledge from the last epoch to soften hard targets in the current epoch \cite{ps-kd}. Memory-replay knowledge distillation (Mr-KD) extends PS-KD by storing a series of abandoned network backups for distillation \cite{Mr-KD}. However, implementing PS-KD or Mr-KD both requires extra GPU memory to store the historical model parameters or the whole past predictions on the disk. The previous strategy is computationally costly for large models such as deep WRN \cite{wrn}, while the latter one is inefficient in training large datasets such as ImageNet \cite{imagenet}. The above-mentioned drawbacks result in low training efficiency, as well as the implementation difficulty on end devices such as mobile phones, digital cameras, etc \cite{Moonshine}, which restrict their applications for regularization. On the other hand, much recent information from the last several mini-batches is absent in these methods. To cope with these shortages, we propose a novel self-distillation framework named DLB, which will be elaborated on in the following section.

%% file: 4-method.tex
\section{Methods}

\subsection{Preliminary}
In this work, we focus on a supervised classification task as case study. For a clear notation, we write a $K$-classes labelled dataset as $\mathcal{D} = \{ (\textbf{x}_i ,y_i ) \}_{i=1}^N$, where $N$ is the total number of training instances. In every mini-batch, a batch of $n$ samples $\mathcal{B} = \{(\textbf{x}_i,y_i)\}_{i=1}^n \subseteq \mathcal{D}$ is augmented by data distortion $\phi$ to derive the distorted images $\mathcal{B}^\phi = \{(\phi(\textbf{x}_i),y_i)\}_{i=1}^n$. Afterwards, they are fed into target neural network $h_\theta$ to optimize the cross-entropy loss function, defined as follows
\begin{equation}
    \mathcal{L}_{CE} = \frac{1}{n}\sum_{i=1}^n H\left(y_i , \textbf{p}_i\right).\label{eq:celoss}
\end{equation}
Formally, the predictive distribution $\textbf{p}_i = (p_{i}(1),\cdots,p_{i}(K))$ in a softmax classifier for class $k\in[K]$ is formulated as
\begin{equation}
    p_{i}(k) = \frac{\exp(f_k(\textbf{x}_i;\theta)/\tau)}{\sum_{j=1}^K \exp(f_j(\textbf{x}_i;\theta)/\tau) },\label{eq:prediction}
\end{equation}
where $f_k$ writes for the $k$\textsuperscript{th} component of the logits from the backbone encoder parameterized by $\theta$. Temperature $\tau$ is usually set to 1 in Eq. \eqref{eq:prediction}. To improve the generalization ability, vanilla knowledge distillation \cite{kd2} transfers pre-trained teacher's knowledge by optimizing an additional Kullback-Leibler (KL) divergence loss between the softened outputs from teacher and student in every mini-batch i.e.
\begin{equation}
    \mathcal{L}_{KD} = \frac{1}{n}\sum_{i=1}^n \tau^2 \cdot D_{KL} \left(\widetilde{\textbf{p}}_i^{\tau}\| \textbf{p}_i^\tau\right),\label{eq:kd}
\end{equation}
where $\textbf{p}_i^\tau$, $\widetilde{\textbf{p}}_i^{\tau}$ are the soften predictions, parameterized by temperature $\tau$, from student and teacher respectively. A higher temperature results in a more uniform distribution, leading to a similar regularization effect as label smoothing \cite{tf-kd,LSR}. Compared with previous works where a complex network is pre-trained to generate $\widetilde{\textbf{p}}_i^{\tau}$, our work uses historic information from the last batch to efficiently generate $\widetilde{\textbf{p}}_i^{\tau}$ as a more instant smoothed labels for regularization.

\subsection{Self-Distillation from Last Batch}
The overall training process for the proposed self-distillation is visualized in \figref{fig:framework}. Instead of adopting a complex pre-trained teacher model to provide sample-wise smoothed labels, our proposed framework utilizes the backup information from the last mini-batch to generate soft targets. It results in a regularization towards training consistency. For a clear notation, we denote the original batch of data sampled in the $t$\textsuperscript{th} iteration as $\mathcal{B}_t = \{ (\textbf{x}_{i}^{t},y_{i}^{t} ) \}_{i=1}^n$, and the network parameters as $\theta_t$. Formally, we substitute the $\widetilde{\textbf{p}}_i^{\tau}$ in Eq. \eqref{eq:kd} by the soften labels $\textbf{p}_{i}^{\tau,t-1}$ generated by the identical network at $t-1$\textsuperscript{th} iteration i.e. $f$ parameterized by $\theta_{t-1}$. Consequently, we introduce an extra last-batch consistency regularization loss to DLB as follows:
\begin{equation}
    \mathcal{L}_{LB} = \frac{1}{n}\sum_{i=1}^n \tau^2 \cdot D_{KL} \left(\textbf{p}_{i}^{\tau,t-1}\| \textbf{p}_{i}^{\tau,t}\right).\label{eq:kd-last-batch}
\end{equation}
Rather than storing the whole $\theta_{t-1}$ in the $t$\textsuperscript{th} iterations as designed in previous works \cite{ps-kd}, which is run-time memory consuming, we complete the computation of all $\textbf{p}_{i}^{\tau,t-1}$ in $t-1$\textsuperscript{th} iteration. We employ a data sampler to obtain $\mathcal{B}_t$ and $\mathcal{B}_{t-1}$ simultaneously at the $t-1$\textsuperscript{th} iteration for implementation. Both the predictions from $\mathcal{B}_{t-1}$ and $\mathcal{B}_{t}$ in $t-1$\textsuperscript{th} iteration update the $\mathcal{L}_{CE}$. Whereas predictions from $\mathcal{B}_t$ are smoothed by temperature $\tau$ and then stored for regularization in $t$\textsuperscript{th} iteration. Storage of a batch of smoothed labels requires very few extra memory cost, which is thus more efficient. Conclusively, the overall loss function is formulated by
\begin{equation}
    \mathcal{L} = \mathcal{L}_{CE} + \alpha \cdot \mathcal{L}_{LB}, \label{eq:overall_loss}
\end{equation}
where $\alpha$ is the coefficient to balance two terms. In short, we constrain half of each mini-batch coinciding with the previous iteration, and the rest half with the upcoming iteration. Afterwards, the former half mini-batch distills on-the-fly soft targets generated in the previous iteration. The overall training process is summarized in Algorithm \ref{alg}.

\begin{algorithm}[htbp!]
\caption{Pseudo code for DLB.} \label{alg}
\begin{algorithmic}[1]
\Require balancing coefficient $\alpha$
\Require distillation temperature $\tau$
\Ensure data\_loader samples batches as in Figure 1
\State last\_logits = None \# initialization
\For{(x,gt\_labels) in data\_loader}
\State $\hat{n}$ = gt\_labels.size(0) \# batch size
\State logits = model.forward(x)
\State loss = CELoss(logits, gt\_labels)
\If{last\_logits != None}
\State soft\_targets = Softmax(last\_logits/$\tau$)
\State pred = Softmax(logits[:$\hat{n}$//2]/$\tau$)
\State loss += $\alpha$*LBLoss(pred, soft\_targets)*$\tau^2$
\EndIf
\State loss.backward() \# update parameters
\State last\_logits = logits[$\hat{n}$//2:].detach() \# no gradient
\EndFor
\end{algorithmic}
\end{algorithm}

%% file: 5-exp.tex
\section{Experiments}

\input{result/r1-table}

\subsection{Datasets and Settings}
\paragraph{Datasets.} We employed three multi-class classification benchmark datasets for comprehensive performance evaluations. The \textbf{CIFAR-10}/\textbf{CIFAR-100} contain a total number of 60,000 RGB natural images of $32\times32$ pixels from 10/100 classes \cite{cifar}. Each class includes 5,000/500 training samples and 1,000/100 testing samples. We followed the widely-used pre-processing from previous works \cite{resnet,wrn}. More precisely, training samples were normalized by deviation and padded 4 pixels with zero-value on each side. A random crop of $32\times32$ region was generated from the padded image or its horizontal flip. The \textbf{TinyImageNet} is a subset of ILSVRC-2012, made up of 200 classes. Each class includes 500 training samples and 50 testing samples, scaled at $64\times64$. All training images were randomly cropped and resized to $32\times32$ after the normalization. The test images were only normalized. 

\paragraph{Implementations.} All experiments were performed on one NVIDIA Tesla V100 GPU with 32Gb memory. The proposed DLB and compared methods were all implemented on Pytorch 1.6.0 in Python 3.7.0 environment. The hyper-parameters for the training scheme followed a consistent setting for a fair comparison. \textbf{CIFAR-100/10}: We followed the settings in previous works \cite{param1,param2}. Specifically, every backbone network was trained for 240 epochs with a batch size of 64. The initial learning rate was set to 0.05, and decayed by a factor of 10\% at 150\textsuperscript{th}, 180\textsuperscript{th} and 210\textsuperscript{th} epoch. We employed a stochastic gradient descent (SGD) optimizer with 0.9 Nesterov momentum, where the weight decay rate was set to $5\times10^{-4}$. \textbf{TinyImageNet}: We also followed the settings in previous works \cite{ols}. Concretely, we set the maximal epoch number to 200, the batch size to 128. The initial learning rate was set to 0.2 with a decayed factor of 10\% at 100\textsuperscript{th}, 150\textsuperscript{th} epoch. The weight decay rate and the momentum in the SGD optimizer were set to $1\times10^{-4}$ and 0.9 respectively. We fixed the temperature $\tau$ at 3 and coefficient $\alpha$ in \name~ to 1, where the dependence of hyper-parameters is explored in the next subsection. The temperature setting in DLB followed suggests from previous work \cite{dml}, and we did not manually tune it for our approach. We used the top-1 error rate (\%) on the test set as the evaluation metric. For reproducibility, we fixed the random seed at 95 in all experiments. We also measured the average and the associated standard deviation over three runs. Importantly, for a fair comparison in terms of computation cost, the proposed method (DLB) was evaluated on half of the total training interactions/epochs than the compared approaches. Although our method comes to a doubled batch size by duplicating half of the last mini-batch, it brings no extra computation cost with halving total training iterations.

\paragraph{Backbone architectures.} We employed six representative architectures \cite{okddip,pcl} for evaluation, namely Vgg-16, Vgg-19 \cite{vgg}, ResNet-32, ResNet-110 \cite{resnet}, WRN20-8 \cite{wrn}, DenseNet-40-12 \cite{densenet}. 

\input{result/r2-table}

\begin{figure*}[!htbp]
\centering
\begin{subfigure}{0.47\linewidth}
\includegraphics[width=1\linewidth]{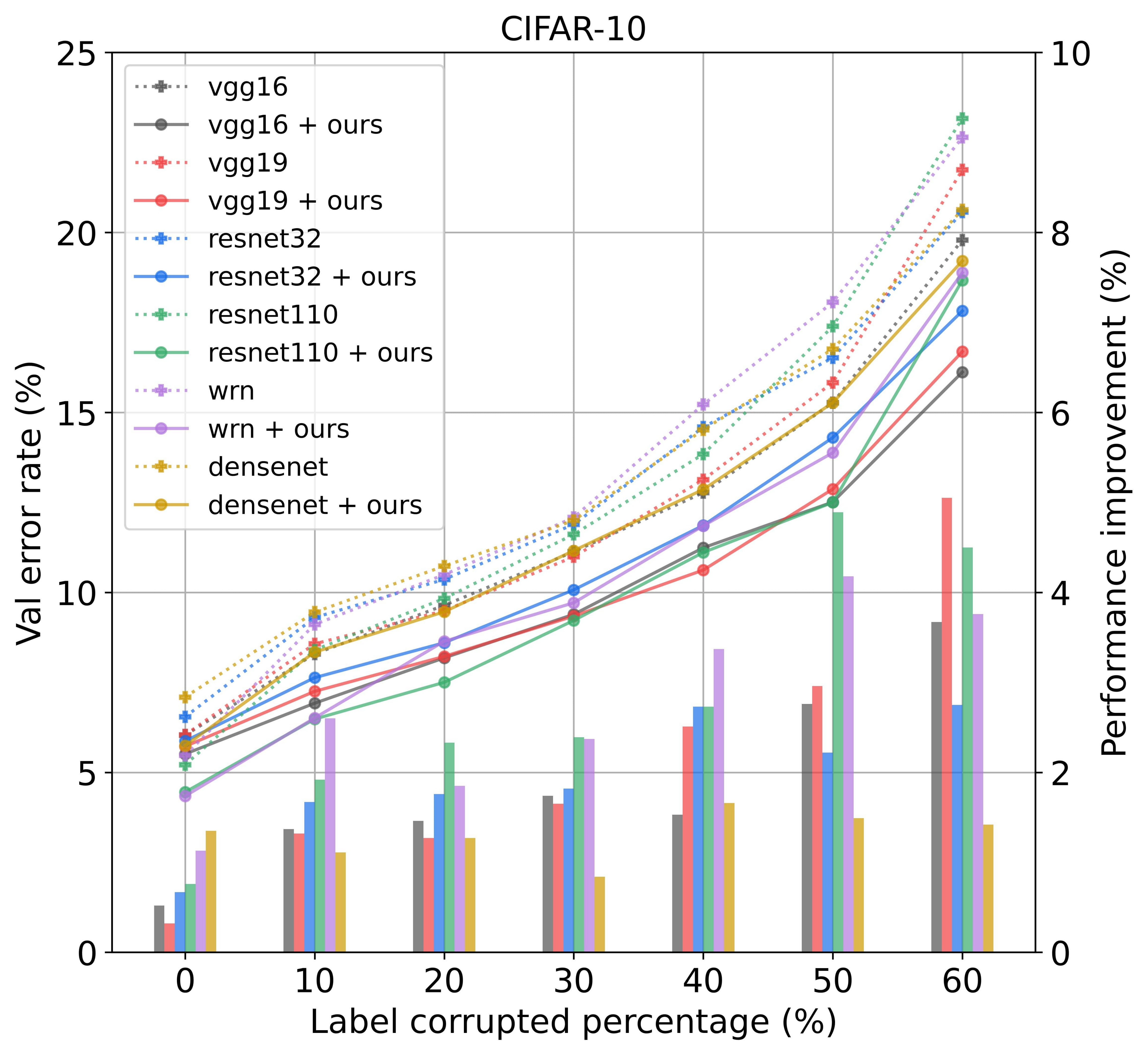}
\caption{CIFAR-100}
\label{fig:result1}
\end{subfigure}
\begin{subfigure}{0.47\linewidth}
\includegraphics[width=1\linewidth]{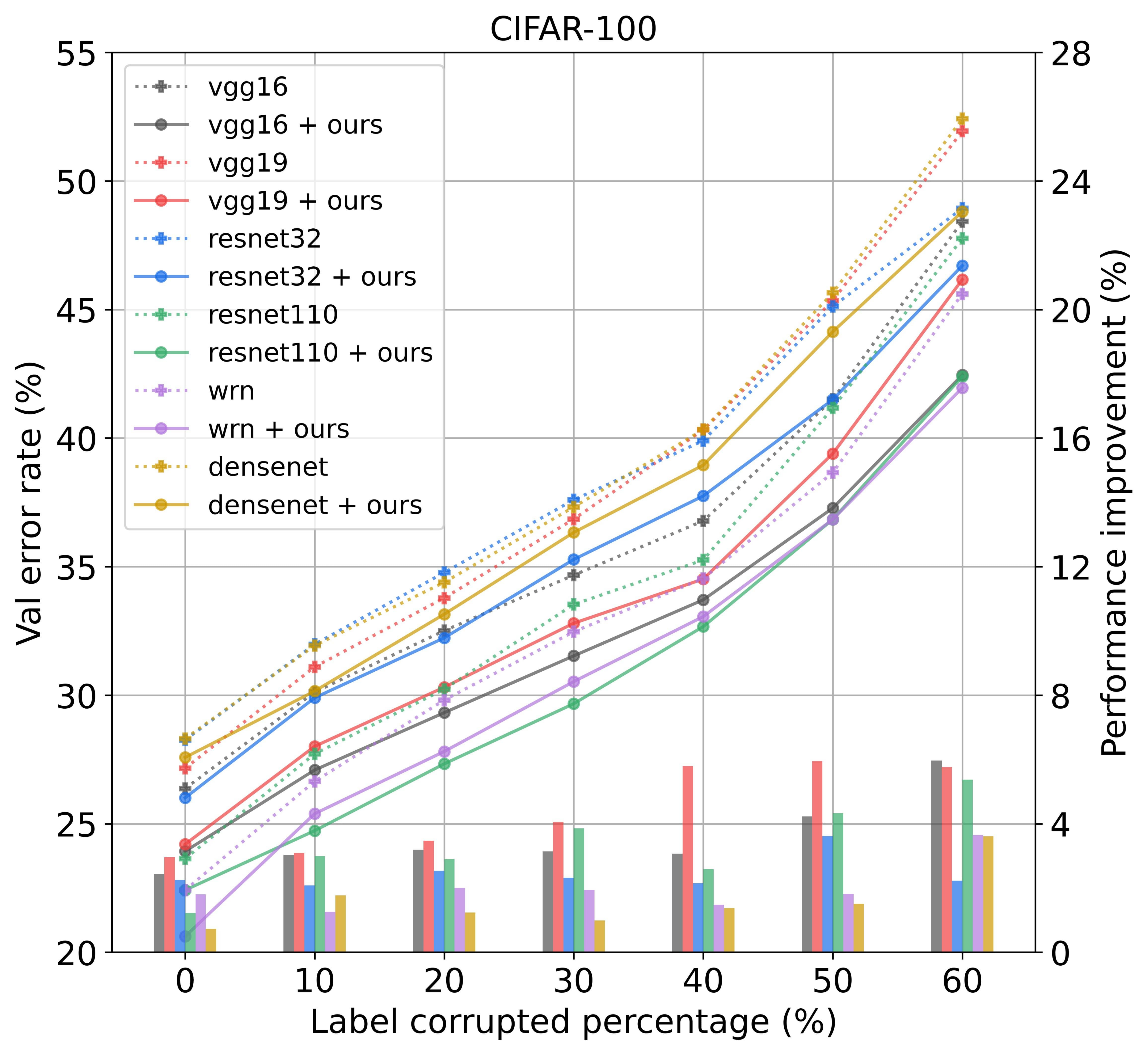}
\caption{CIFAR-10}
\label{fig:result2}
\end{subfigure}
\caption{Performance on training with label noise. (a)  CIFAR-100. (b) CIFAR-10. The line shows the top-1 validation error rate (\%) on corrupted data at different percentage ($p$) of the noisy label i.e., $p\in\{0,10\%,20\%,30\%,40\%,50\%,60\%\}$. The bar illustrates the improvement of our approach, compared with baseline. 
}
\label{fig:result_pic}
\end{figure*}

\begin{figure*}[!htbp]
\centering
\begin{subfigure}{0.23\linewidth}
\includegraphics[width=1\linewidth]{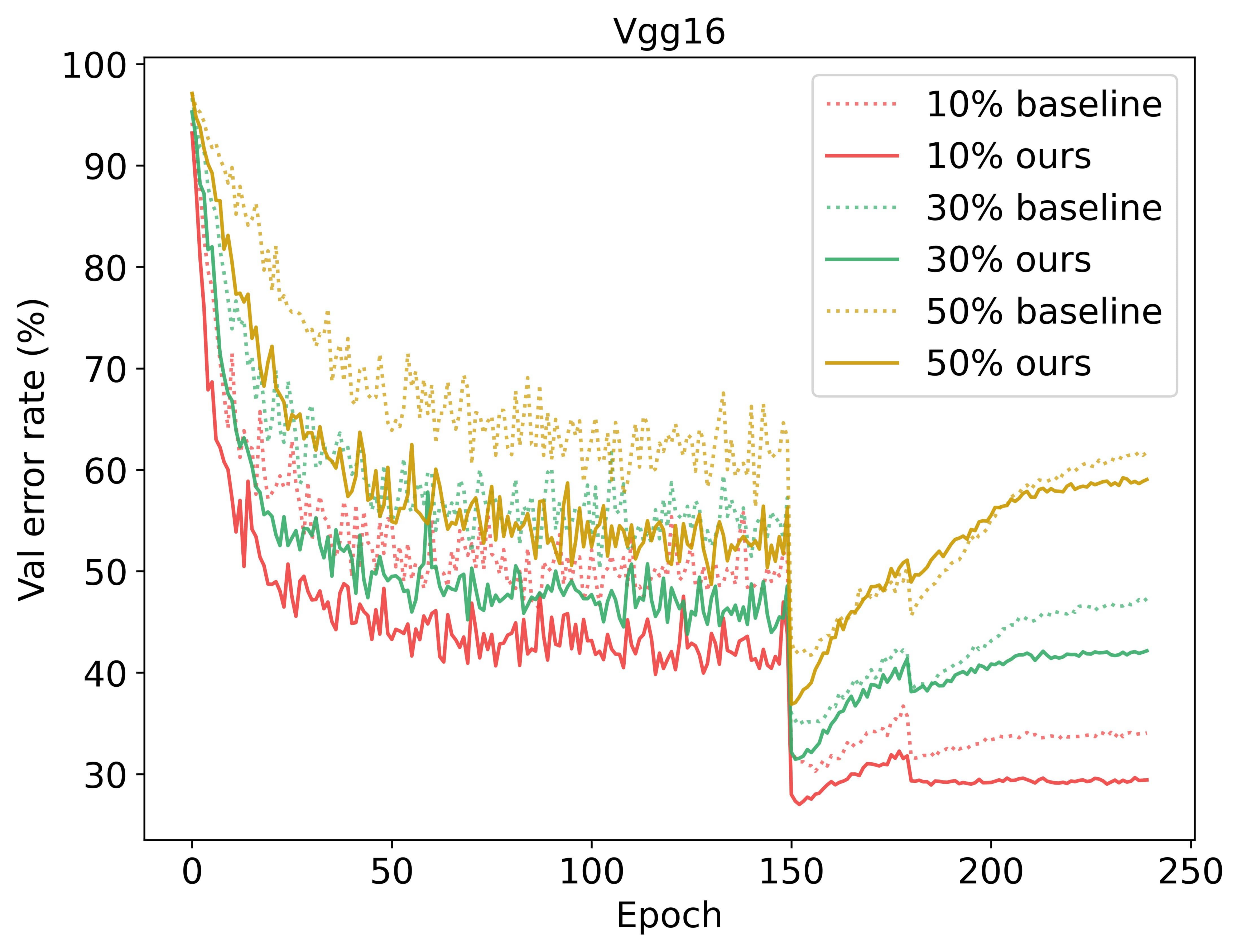}
\caption{Vgg-16}
\label{fig:result_sup_1}
\end{subfigure}
%
%
\begin{subfigure}{0.23\linewidth}
\includegraphics[width=1\linewidth]{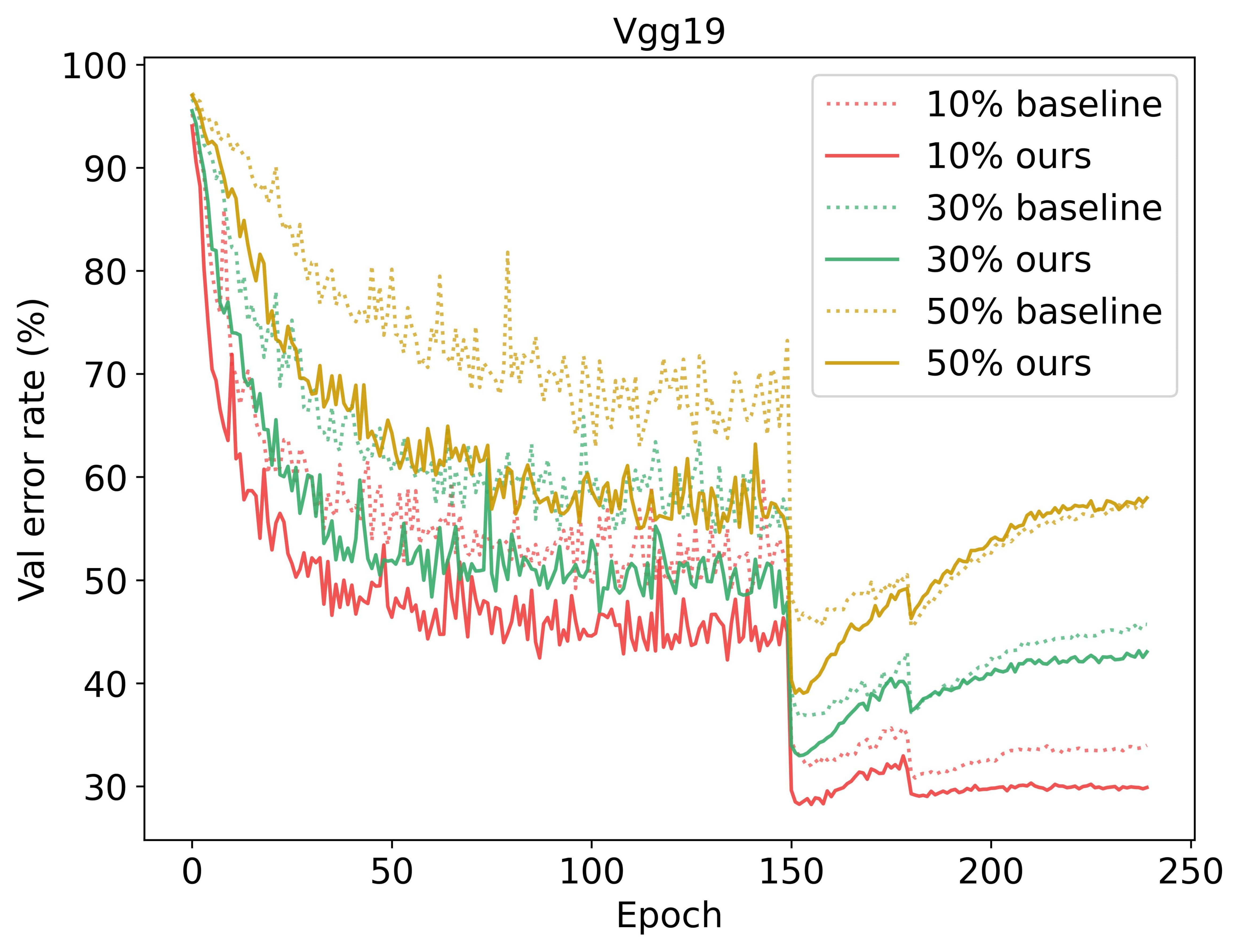}
\caption{Vgg-19}
\label{fig:result_sup_2}
\end{subfigure}
%
%
\begin{subfigure}{0.23\linewidth}
\includegraphics[width=1\linewidth]{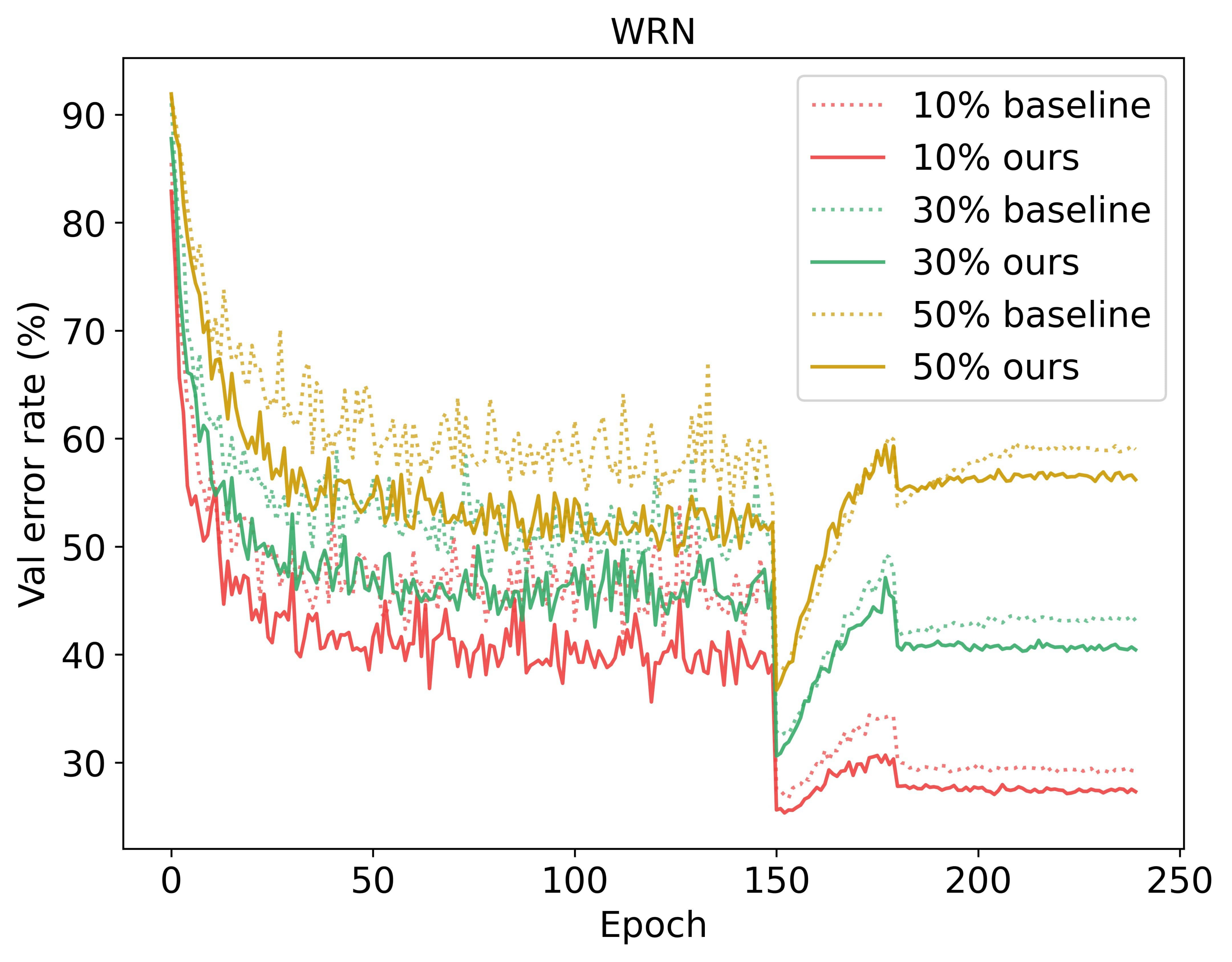}
\caption{WRM}
\label{fig:result_sup_3}
\end{subfigure}
\begin{subfigure}{0.23\linewidth}
\includegraphics[width=1\linewidth]{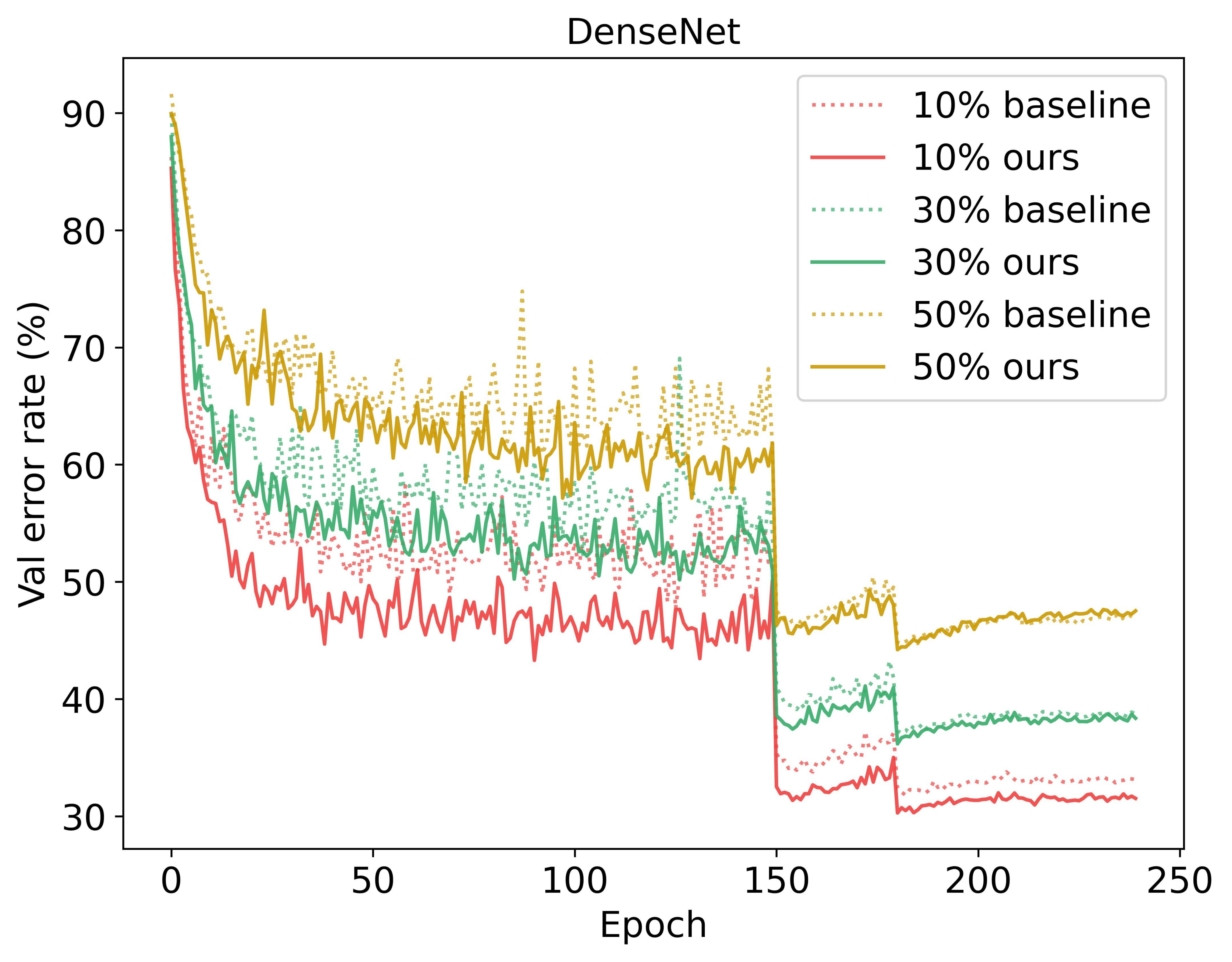}
\caption{DenseNet}
\label{fig:result_sup_4}
\end{subfigure}
\\
\begin{subfigure}{0.23\linewidth}
\includegraphics[width=1\linewidth]{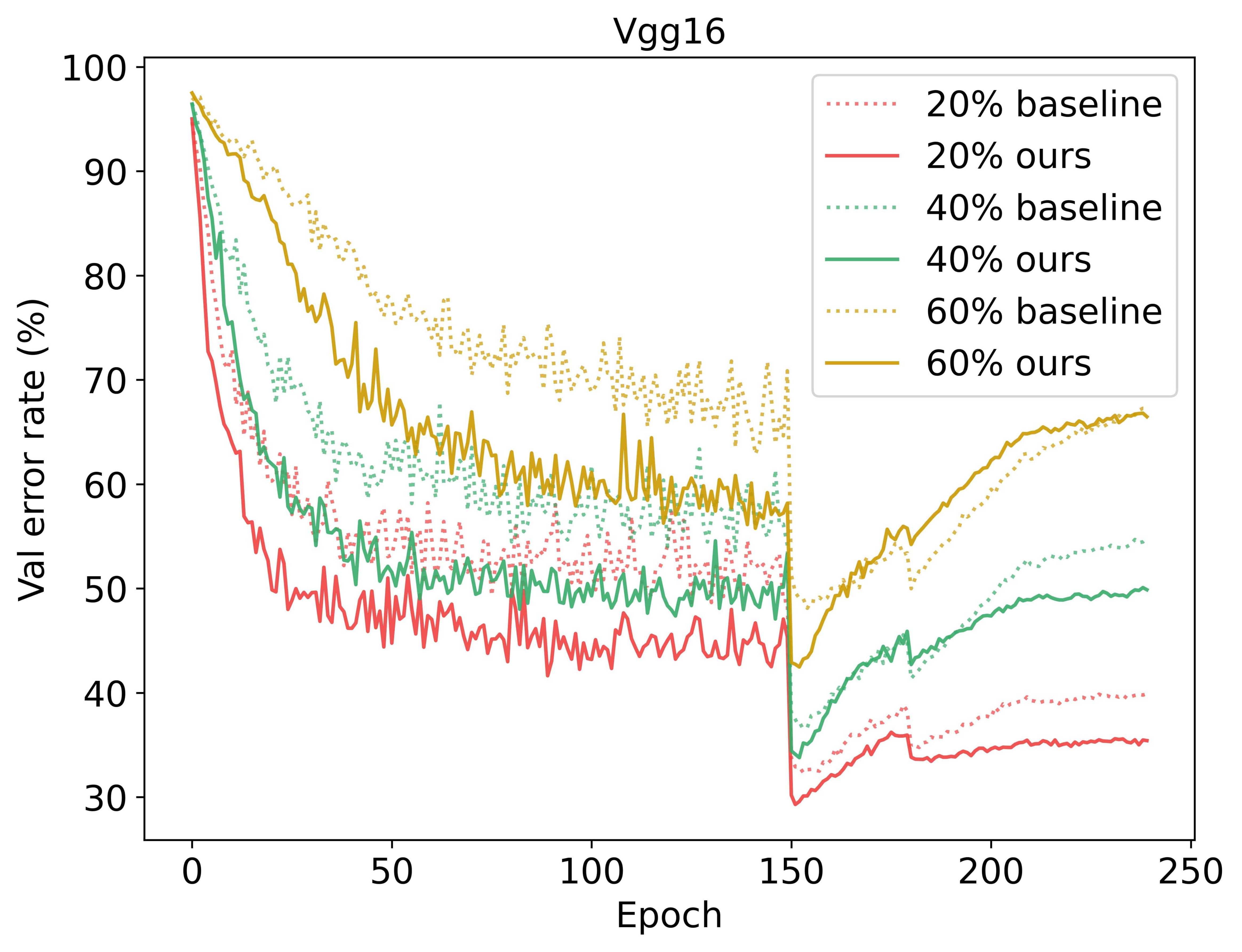}
\caption{Vgg-16}
\label{fig:result_sup_12}
\end{subfigure}
\begin{subfigure}{0.23\linewidth}
\includegraphics[width=1\linewidth]{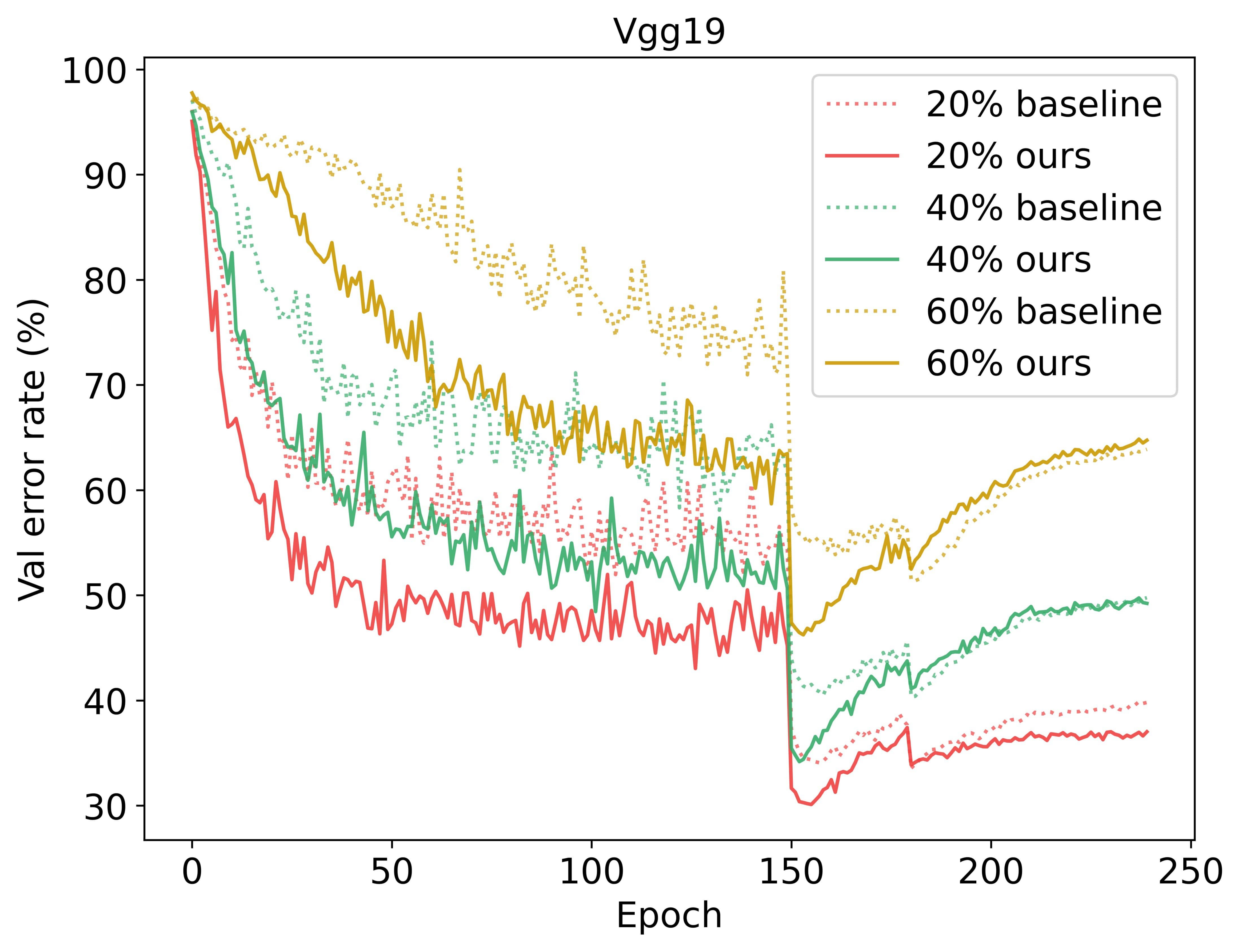}
\caption{Vgg-19}
\label{fig:result_sup_22}
\end{subfigure}
\begin{subfigure}{0.23\linewidth}
\includegraphics[width=1\linewidth]{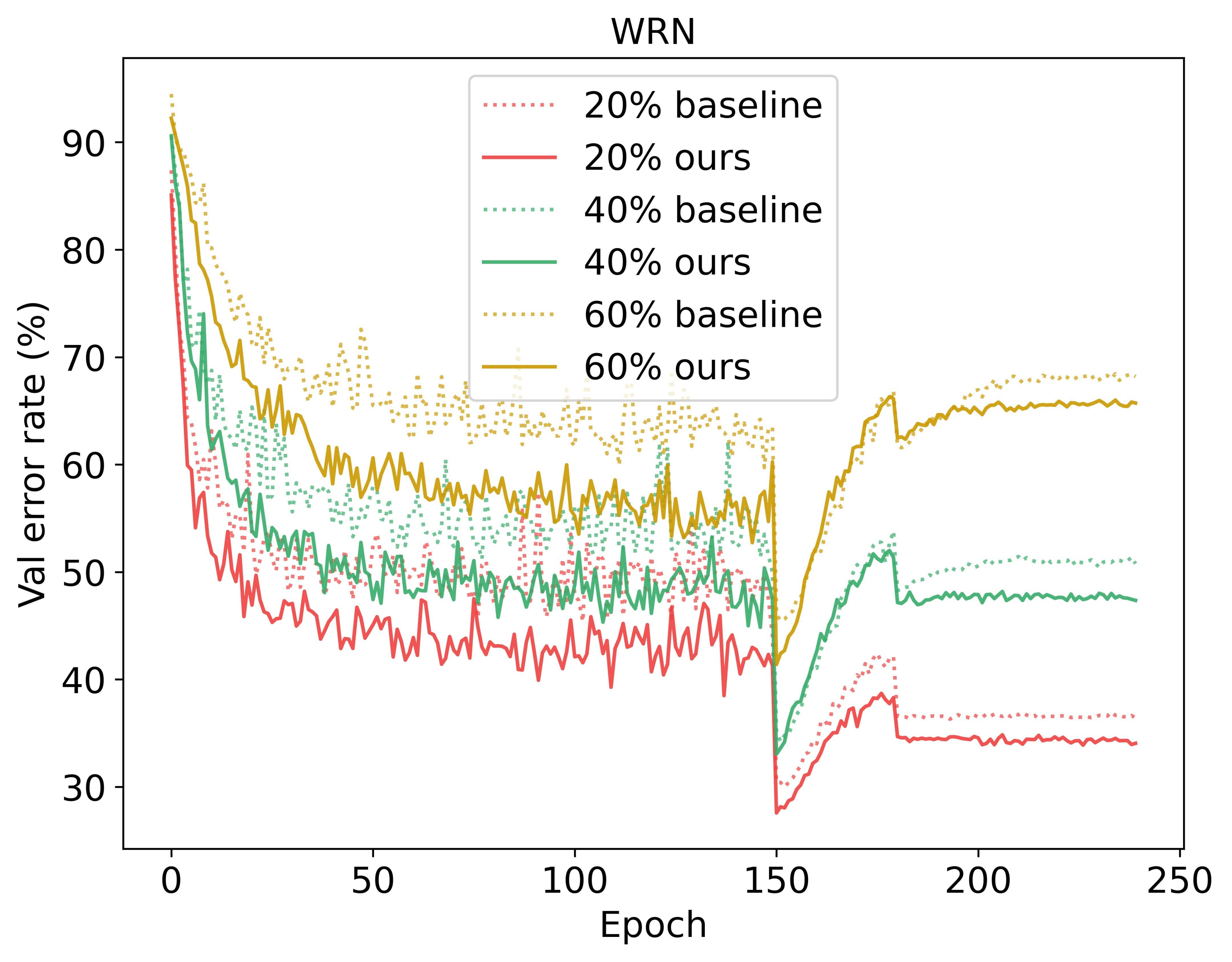}
\caption{WRM}
\label{fig:result_sup_32}
\end{subfigure}
\begin{subfigure}{0.23\linewidth}
\includegraphics[width=1\linewidth]{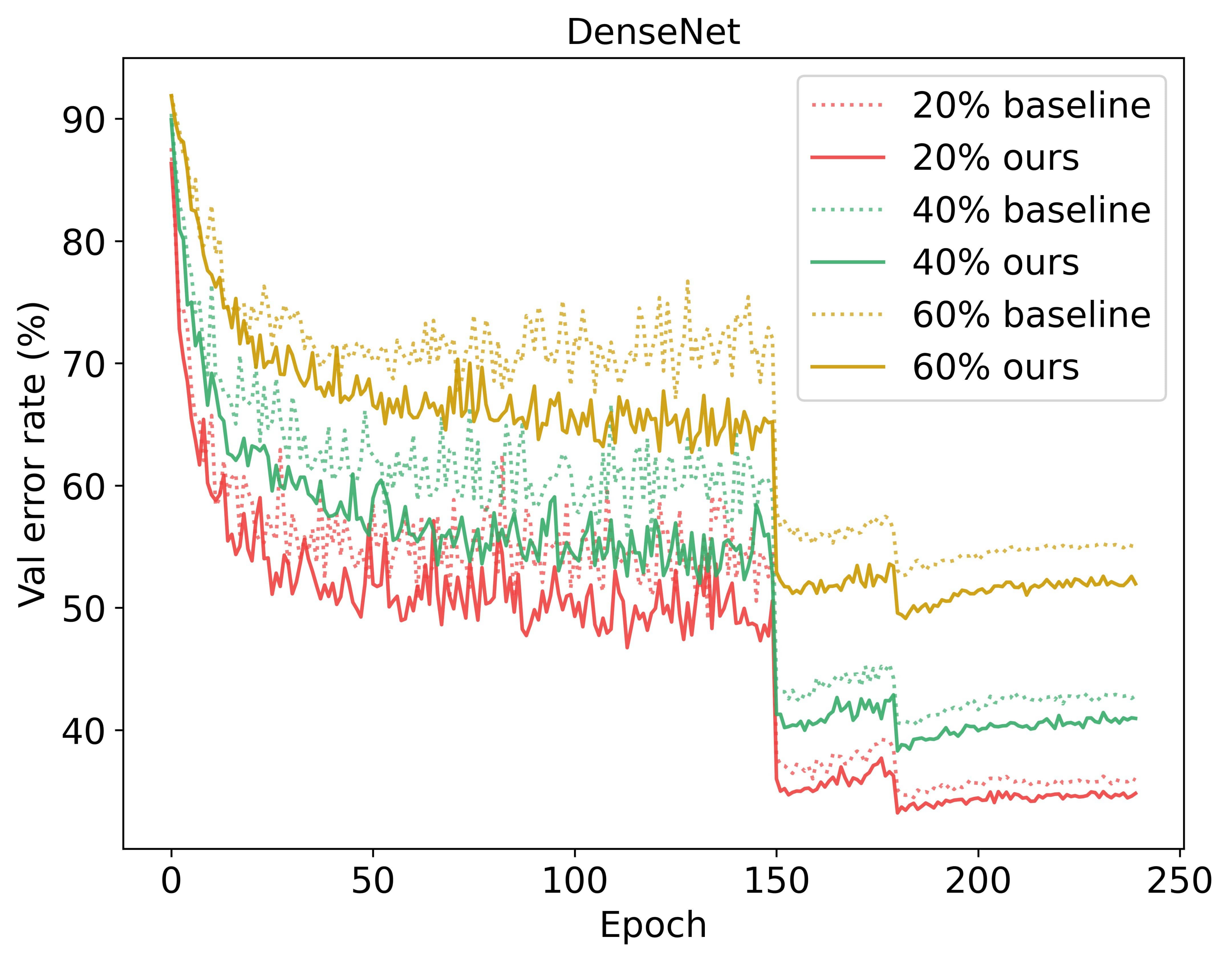}
\caption{DenseNet}
\label{fig:result_sup_42}
\end{subfigure}
%
\caption{The training performance of Vgg-16, Vgg-19, WRN, DenseNet on corrupted data. (a)-(d) Results on data corruption rate $p\in\{0.1,0.3,0.5\}$; (e)-(h) Results on data corruption rate $p\in\{0.2,0.4,0.6\}$.  }
\label{fig:result_sup}
\end{figure*}

\begin{figure}[!b]

\centering
\begin{subfigure}{0.48\linewidth}
\includegraphics[width=1\linewidth]{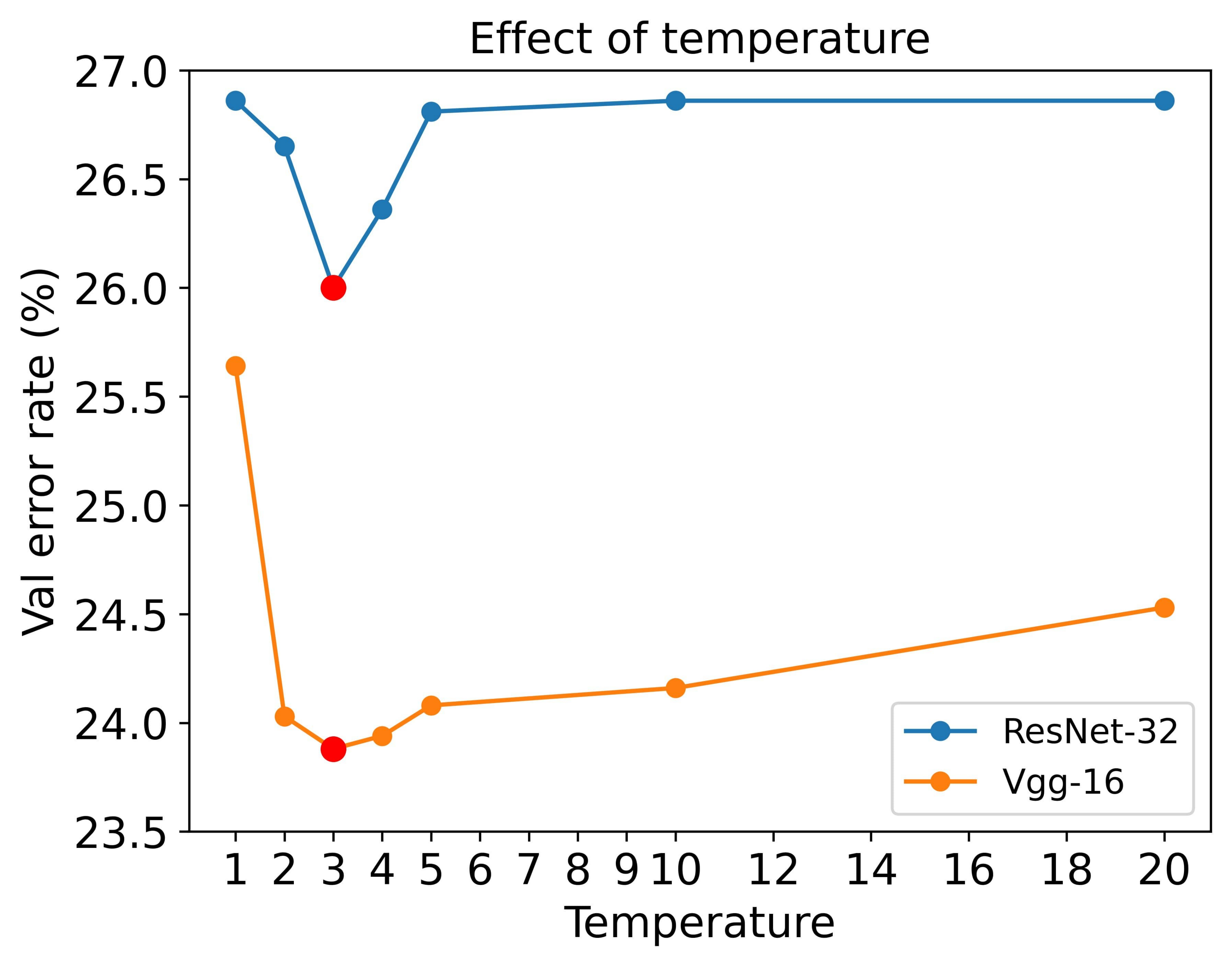} 
\caption{Temperature}
\label{fig:result_ab_t}
\end{subfigure}
\begin{subfigure}{0.48\linewidth}
\includegraphics[width=1\linewidth]{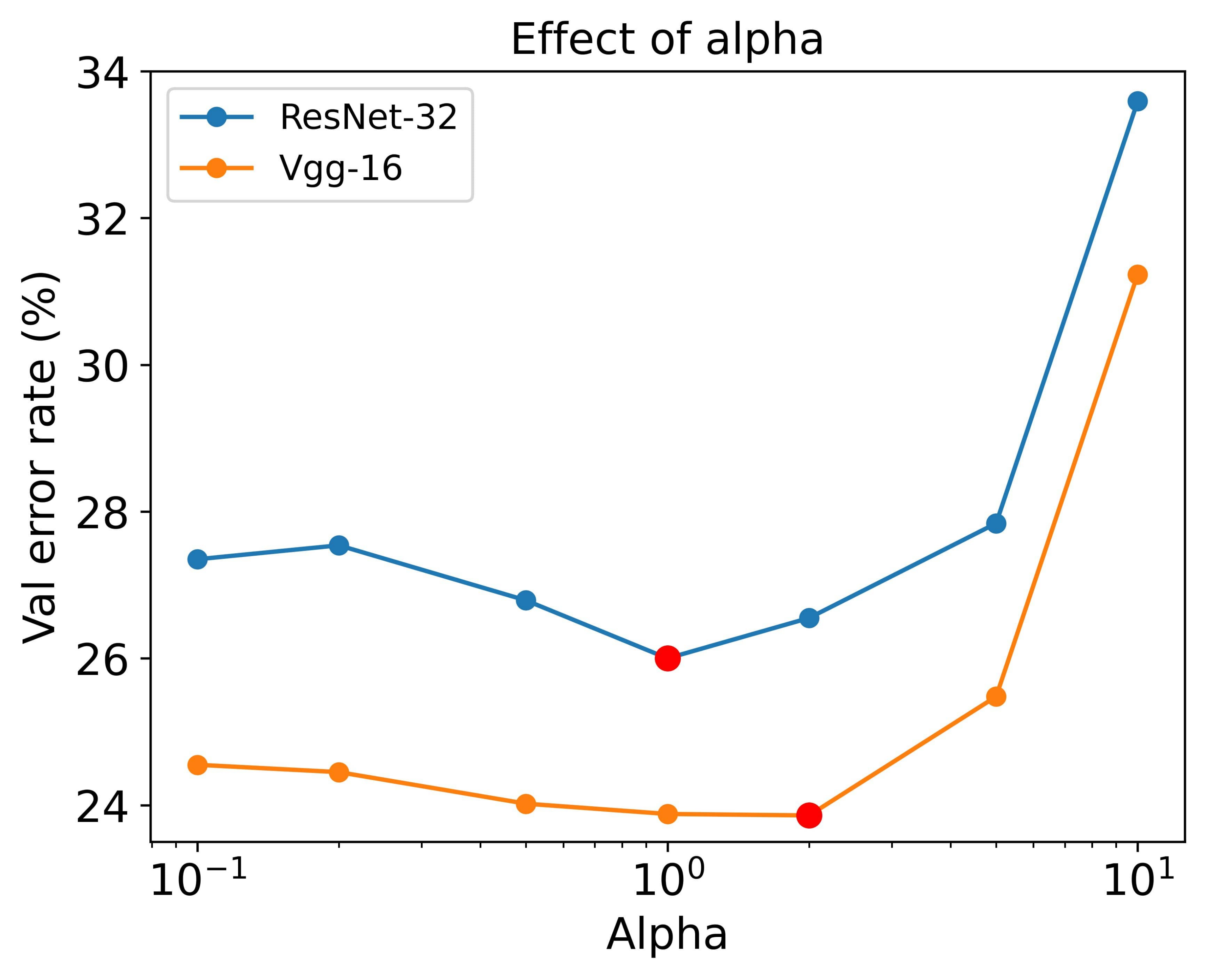} 
\caption{Alpha}
\label{fig:result_ab_a}
\end{subfigure}
\caption{Impact of hyper-parameters in \name~ on CIFAR-100 to ResNet-32 and Vgg-16. (a) Impact of temperature $\tau \in \{1,2,3,4,5,10,20\}$, with a fixed $\alpha=1$. (b) Impact of balancing coefficient $\alpha\in\{0.1,0.2,0.5,1.0,2.0,5.0,10.0\}$ with fixed temperature 3. The best performance in each controlled experiment is marked out by red.}
\label{fig:ablation}
\end{figure}

\paragraph{Compared methods.} Hard labels were utilized in the baseline to train the target network directly. We also compared the proposed method with label smoothing regularization (LSR) \cite{LSR} and self-knowledge distillation regularization approaches, including teacher-free knowledge distillation (Tf-KD$_{self}$,Tf-KD$_{reg}$) \cite{tf-kd}, class-wise self-knowledge distillation (CS-KD) \cite{cs-kd}, progressive self-knowledge distillation (PS-KD) \cite{ps-kd}. The above methods focus on logit-level regularization. Data-distortion guided self-knowledge distillation (DDGSD) is a data augmentation based distillation approach \cite{r-aug}, which was compared and tested the compatibility with DLB. We removed the feature-level supervision in DDGSD \cite{r-aug} i.e. MMD loss, for a fair comparison. As DDGSD is an augmentation based method, we explore the performance compatibility between \name~ and DDGSD. All the extra hyper-parameters involved in the compared methods were retained as their original settings. 

\subsection{Classification Results}
As illustrated in \tabref{table:result_cifar}, our method (denoted as \name) consistently improved the performance on various backbones (baseline). To be more specific, the averaged error rate improvement achieved by \name~ ranged from 0.83\% to 2.50\% on CIAFR-100, 0.37\% to 1.01\% on CIFAR-10, and 0.81\% to 3.17 on TinyImageNet. It shows the effectiveness of DLB that can significantly improve the generalization ability on various classification tasks. Moreover, DLB outperformed the state-of-the-art approaches, achieving the lowest top-1 error. The best and second-best performances on each set were highlighted in red and green respectively. We can observe that DLB on CIFAR-10 succeeded the state-of-the-arts by 0.30\% with WRN20-8; whilst on CIFAR-100 by 0.47\%. These improvements are attributed to the self-distillation regularization from the last mini-batch. We notice that DLB significantly outperformed Tf-KD \cite{tf-kd}, and PS-KD \cite{ps-kd}. It demonstrates the performance advantage of DLB to generalize CNN. Furthermore, the identical teacher from the last mini-batch provides dynamically updated smoothed labels that fit the training process better than a pre-trained teacher or the last-epoch backup. Conclusively, \name~ can efficiently serve as a universal regularization to normally train neural networks.

\subsection{Compatibility with Augmentations}
Previous work claims that data augmentation and distillation provide an orthogonal improvement \cite{combine_aug}. To evaluate the compatibility of DLB with data augmentation based regularization, we combined our method with CutMix\cite{cutmix}, CutOut\cite{cutout} and DDGSD \cite{r-aug} on CIFAR-10/100.

\paragraph{CutOut.} CutOut randomly masks a square region in the training samples \cite{cutout}, we set the number of holes to 1 and the hole size to 16. Combining DLB with CutOut is straightforward. As shown in \tabref{table:result_cifar_augmentation}, DLB can progressively improve CutOut by 0.45\% to 1.14\% on CIFAR-10 and 0.84\% to 2.74\% on CIFAR-100. The performance enhancement was slightly higher than itself on the baseline. It suggests that DLB works in conjunction with CutOut. 

\paragraph{CutMix.} CutMix randomly cuts and pasts patches in a mini-batch for regularization, where we follow the additional hyper-parameters setting as its original work \cite{cutmix} i.e., $\beta=1$ for beta distribution and augmentation probability $p=0.5$. CutMix improved the baseline by approximately 0.52\% on CIFAR-10 and 1.48\% on CIFAR-100. We plugged CutMix into our approach by performing self-distillation on a batch of images distorted by the same CutMix operation. It resulted in an extra 0.42\% improvement on CIFAR-10 and 1.02\% on CIFAR-100.

\paragraph{DDGSD.} DDGSD is a self-distillation scheme, excavating a consistency between different distorted versions of the same images \cite{r-aug}. We plugged \name~ into DDGSD by distilling the last mini-batch for both two predictions from different augmented versions. As demonstrated in \tabref{table:result_cifar_augmentation}, DLB slightly outperforms DDSGD. Moreover, combining DLB with DDSGD can obtain observable improvements.

\paragraph{Discussion.} We empirically show that DLB and augmentation regularization work orthogonally. To be more specific, an extra performance gain can be achieved by combing DLB with augmentation based regularization. These finds indicate that using \name~ as a plugged-in regularization can enhance the generation of other approaches.

\subsection{Robustness to Data Corruption}
In this section, we try to understand how \name~ works by empirically exploring its regularization effect on a corrupted setting. DLB is expected to impose training stability and consistency. The superiority of DLB is its simplicity for implementation and parallelization. Additionally, \name~ is expected to amplify the robustness of the target neural network, especially in noisy data. It results in a stronger tolerance to the corrupted data, by mitigating the over-fitting to label noise \cite{dnn_robustness}. To prove this statement, we kept the experimental settings aligned with previous works \cite{ols,label_noise,label_noise2} by randomly injecting symmetric label noise at different rate $p\in\{0.1,0.2,0.3,0.4,0.5,0.6\}$ to CIFAR-100/CIFAR-10 before the training process. Whilst, the test set was kept clean. In \figref{fig:result_pic} (a)-(b), we can observe a stable performance improvement on different models. For example, on CIFAR-100, DLB improved baseline by 2.38\% at $p=0.1$, and 4.44\% at $p=0.6$. The generalization improvement went increasingly higher with a higher label noise rate $p$. These observations demonstrate that \name~ can effectively mitigate a neural network to fit label noise and improve the overall performance. The training performance of Vgg-16, Vgg-19, WideResNet, DenseNet are visualized in \figref{fig:result_sup}.

\subsection{Impact of Hyper-parameters}
To analyze the impact of two hyper-parameters in \name, namely the temperature $\tau$ in Eq. \eqref{eq:kd-last-batch} and balancing coefficient $\alpha$ in Eq. \eqref{eq:overall_loss}, we carried out controlled experiments. Firstly, we fixed $\alpha$ to 1 and assigned $\tau$ with different values ranged in $\{1,2,3,4,5,10,20\}$ to evaluate the performance of ResNet-32, Vgg-16 on CIFAR-100. As plotted in \figref{fig:ablation} (a), \name~ achieved lowest error rate with a temperature $\tau = 3$. The performance reliance on $\alpha$ is demonstrated in \figref{fig:ablation} (b), with a fixed $\tau =3$. We observe that \name~ performs well when $\alpha$ changes in $[0.5,2.0]$.

\subsection{Ablation Study}
We removed the last batch distillation loss and only retained the cross-entropy loss for ablation on CIFAR-100. Specifically, ResNet-32 achieved a $28.01$\% test top-1 error rate, which performs slightly better than the baseline ($28.26$\%). However, it was far worse than our \name~($26.00$\%). These findings show the significance to include the distillation loss as well as the effectiveness of our method. Detailed ablation results on all models are reported in the \tabref{table:ablation}. 

\input{result/r3-table}

%% file: result/r1-table.tex
\begin{table*}[!ht]
\caption{
Performance comparison with label smoothing regularization and state-of-the-art self-distillation methods on CIFAR-10, CIFAR-100 and TinyImageNet in terms of average top-1 error rate (\%). We calculated the average and deviation over three runs. The best and second best performance were highlighted in \textcolor{myRed}{\textbf{Red}} and \textcolor{myGreen}{Green} respectively.
}\label{table:result_cifar}
\centering
\begin{tabular}{c|l|cccccc} 
\toprule
Dataset&Methods & Vgg-16 & Vgg-19 & ResNet-32 & ResNet-110 & WRN20-8 & DenseNet-40-12\\
\hline 
\multirow{8}{*}{CIFAR-10}
& Baseline & 6.03\tiny{$\pm$0.22} & 6.04\tiny{$\pm$0.10} & 6.54\tiny{$\pm$0.10} & 5.21\tiny{$\pm$0.24} & 5.47\tiny{$\pm$0.08} & 7.09\tiny{$\pm$0.28}\\
& LSR & 5.91\tiny{$\pm$0.40} & 6.05\tiny{$\pm$0.06} & 6.73\tiny{$\pm$0.17} & 5.60\tiny{$\pm$0.17} & \textcolor{myGreen}{4.76\tiny{$\pm$0.06}} & 7.47\tiny{$\pm$0.27} \\
& Tf-KD$_{self}$ & 5.92\tiny{$\pm$0.15} & \textcolor{myGreen}{5.91\tiny{$\pm$0.01}} & 6.32\tiny{$\pm$0.01} & \textcolor{myGreen}{4.92\tiny{$\pm$0.08}} & 5.33\tiny{$\pm$0.13} & 7.03\tiny{$\pm$0.14}\\
& Tf-KD$_{reg}$ & 6.12\tiny{$\pm$0.07} & 6.25\tiny{$\pm$0.18} & 6.60\tiny{$\pm$0.05} & 5.48\tiny{$\pm$0.18} & 4.79\tiny{$\pm$0.01} & 7.38\tiny{$\pm$0.16}\\
& CS-KD & 6.22\tiny{$\pm$0.20} & 6.38\tiny{$\pm$0.11} & 6.88\tiny{$\pm$0.15} & 6.12\tiny{$\pm$0.05} & 4.89\tiny{$\pm$0.19} & 7.85\tiny{$\pm$0.17}\\
& PS-KD & \textcolor{myGreen}{5.90\tiny{$\pm$0.04}} & 6.07\tiny{$\pm$0.57} & \textcolor{myGreen}{5.96\tiny{$\pm$0.06}} & 5.09\tiny{$\pm$0.68} & 4.99\tiny{$\pm$0.01} & \textcolor{myGreen}{6.77\tiny{$\pm$0.28}}\\
\cline{2-8}
& \multirow{2}{*}{\name} & \textcolor{myRed}{\textbf{5.38}\tiny{$\pm$0.01}} & \textcolor{myRed}{\textbf{5.58}\tiny{$\pm$0.08}} & \textcolor{myRed}{\textbf{5.85}\tiny{$\pm$0.02}} & \textcolor{myRed}{\textbf{4.85}\tiny{$\pm$0.11}} & \textcolor{myRed}{\textbf{4.46}\tiny{$\pm$0.01}} & \textcolor{myRed}{\textbf{6.57}\tiny{$\pm$0.02}}  \\
& & \small(0.65 \upmark) & \small(0.46 \upmark) & \small(0.70 \upmark) & \small(0.37 \upmark) & \small(1.01 \upmark) & \small(0.53 \upmark) \\
\midrule
\multirow{8}{*}{CIFAR-100}
& Baseline & 26.37\tiny{$\pm$0.19} & 27.16\tiny{$\pm$0.42} & 28.26\tiny{$\pm$0.18} & 23.64\tiny{$\pm$0.25} & 22.42\tiny{$\pm$0.23} & \textcolor{myGreen}{28.31\tiny{$\pm$0.35}}\\
& LSR & \textcolor{myGreen}{25.81\tiny{$\pm$0.02}} & 26.75\tiny{$\pm$0.45} & 28.21\tiny{$\pm$0.13} & 23.32\tiny{$\pm$0.16} & 22.17\tiny{$\pm$0.01} & 29.07\tiny{$\pm$0.01}\\
& Tf-KD$_{self}$ & 25.94\tiny{$\pm$0.11} & 27.46\tiny{$\pm$0.82} & \textcolor{myGreen}{26.09\tiny{$\pm$0.24}} & 27.02\tiny{$\pm$0.01} & 21.88\tiny{$\pm$0.34} & 28.40\tiny{$\pm$0.09}\\
& Tf-KD$_{reg}$ & 25.85\tiny{$\pm$0.30} & 26.82\tiny{$\pm$0.59} & 28.29\tiny{$\pm$0.07} & 23.54\tiny{$\pm$0.04} & 22.28\tiny{$\pm$0.05} & 28.92\tiny{$\pm$0.04}\\
& CS-KD & 25.81\tiny{$\pm$0.43} & 26.65\tiny{$\pm$0.49} & 29.21\tiny{$\pm$0.21} & 23.41\tiny{$\pm$0.28} & 21.75\tiny{$\pm$0.26} &
29.65\tiny{$\pm$0.51} \\
& PS-KD & 25.95\tiny{$\pm$0.74} & \textcolor{myGreen}{26.36\tiny{$\pm$0.76}} & 27.49\tiny{$\pm$0.75} & \textcolor{myGreen}{22.85\tiny{$\pm$0.65}} & \textcolor{myGreen}{21.26\tiny{$\pm$0.11}} & 28.48\tiny{$\pm$0.53}\\
\cline{2-8}
& \multirow{2}{*}{\name} & \textcolor{myRed}{\textbf{23.88}\tiny{$\pm$0.06}} & \textcolor{myRed}{\textbf{24.53}\tiny{$\pm$0.13}} & \textcolor{myRed}{\textbf{26.00}\tiny{$\pm$0.03}} & \textcolor{myRed}{\textbf{21.82}\tiny{$\pm$0.19}} & \textcolor{myRed}{\textbf{20.79}\tiny{$\pm$0.33}} & \textcolor{myRed}{\textbf{27.48}\tiny{$\pm$0.21}} \\
& & \small(2.50 \upmark) & \small(2.63 \upmark) & \small(2.26 \upmark) & \small(1.83 \upmark) & \small(1.63 \upmark) & \small(0.83 \upmark) \\
\midrule
\multirow{8}{*}{TinyImageNet}
& Baseline & 48.83\tiny{$\pm$0.33} & 50.02\tiny{$\pm$0.12} & 50.38\tiny{$\pm$0.38} & 43.20\tiny{$\pm$0.56} & 43.72 \tiny{$\pm$0.28} & 50.94\tiny{$\pm$0.56}\\
& LSR & 47.95\tiny{$\pm$0.32} & 48.86\tiny{$\pm$0.78} & 52.75\tiny{$\pm$0.69} & 42.18\tiny{$\pm$0.57} & 43.51\tiny{$\pm$0.13} & 51.01\tiny{$\pm$0.16} \\
& Tf-KD$_{self}$ & \textcolor{myGreen}{46.76\tiny{$\pm$0.10}} & \textcolor{myGreen}{47.25\tiny{$\pm$0.06}} & \textcolor{myGreen}{49.48\tiny{$\pm$0.16}} & \textcolor{myGreen}{41.31\tiny{$\pm$0.01}} & \textcolor{myGreen}{41.13\tiny{$\pm$0.11}} & \textcolor{myGreen}{50.78\tiny{$\pm$0.01}} \\
& Tf-KD$_{reg}$ & 48.31\tiny{$\pm$0.06} & 49.54\tiny{$\pm$0.16} & 50.97\tiny{$\pm$0.02} & 41.88\tiny{$\pm$0.91} & 42.85\tiny{$\pm$0.07} & 51.47\tiny{$\pm$0.10}\\
& CS-KD & 46.95\tiny{$\pm$0.15} & 47.89\tiny{$\pm$0.13} & 53.99\tiny{$\pm$0.03} & 43.06\tiny{$\pm$0.22} & 42.04\tiny{$\pm$0.42} & 54.93\tiny{$\pm$0.25} \\
& PS-KD & 48.39\tiny{$\pm$0.23} & 49.77\tiny{$\pm$0.23} & 50.28\tiny{$\pm$0.17} & 42.22\tiny{$\pm$0.72} & 43.37\tiny{$\pm$0.01} & 51.57\tiny{$\pm$0.05} \\
\cline{2-8}
& \multirow{2}{*}{\name} & \textcolor{myRed}{\textbf{45.66}\tiny{$\pm$0.01}} & \textcolor{myRed}{\textbf{46.68}\tiny{$\pm$0.09}} & \textcolor{myRed}{\textbf{48.66}\tiny{$\pm$0.07}} & \textcolor{myRed}{\textbf{40.39}\tiny{$\pm$0.01}} & \textcolor{myRed}{\textbf{41.03}\tiny{$\pm$0.02}} & \textcolor{myRed}{\textbf{50.13}\tiny{$\pm$0.01}} \\
& & \small(3.17 \upmark) & \small(3.35 \upmark) & \small(1.72 \upmark) & \small(2.81 \upmark) & \small(2.69 \upmark) & \small(0.81 \upmark) \\
\bottomrule
\end{tabular}
\end{table*}

%% file: result/r2-table.tex
\begin{table*}[!ht]
\caption{
Performance compatibility with augmentation-based regularization methods including CutOut \cite{cutout}, CutMix \cite{cutmix} and DDGSD \cite{r-aug} on CIFAR-10/100. We calculated the average top-1 error rate (\%), standard deviation of three runs, written in the form of 'avg $\pm$ std'. The best result in each category was highlighted in \textbf{boldface}.
}\label{table:result_cifar_augmentation}
\centering
\resizebox{\linewidth}{!}{
\begin{tabular}{c|l|cccccc} 
\toprule
Dataset&Methods & Vgg-16 & Vgg-19 & ResNet-32 & ResNet-110 & WRN20-8 & DenseNet-40-12\\
\hline 
\multirow{8}{*}{C10}
& Baseline & 6.03\tiny{$\pm$0.22} & 6.04\tiny{$\pm$0.10} & 6.54\tiny{$\pm$0.10} & 5.21\tiny{$\pm$0.24} & 5.47\tiny{$\pm$0.08} & 7.09\tiny{$\pm$0.28}\\
& + \name & \textbf{5.38}\tiny{$\pm$0.01} & \textbf{5.58}\tiny{$\pm$0.08} & \textbf{5.85}\tiny{$\pm$0.02} & \textbf{4.85}\tiny{$\pm$0.11} & \textbf{4.46}\tiny{$\pm$0.01} & \textbf{6.57}\tiny{$\pm$0.02} \\
\cline{2-8}
& + CutOut & 4.93\tiny{$\pm$0.04} & 5.10\tiny{$\pm$0.05} & 5.91\tiny{$\pm$0.12} & 4.64\tiny{$\pm$0.14} & 4.88\tiny{$\pm$0.11} & 6.68\tiny{$\pm$0.11} \\
& + CutOut + \name & \textbf{4.48}\tiny{$\pm$0.03} & \textbf{4.46}\tiny{$\pm$0.40} & \textbf{5.18}\tiny{$\pm$0.05} & \textbf{3.78}\tiny{$\pm$0.09} & \textbf{4.12}\tiny{$\pm$0.09} & \textbf{5.54}\tiny{$\pm$0.04} \\
\cline{2-8}
& + CutMix & 5.29\tiny{$\pm$0.04} & 5.41\tiny{$\pm$0.03} & 5.96\tiny{$\pm$0.28} & 4.82\tiny{$\pm$0.10} & 4.89\tiny{$\pm$0.31} & 6.76\tiny{$\pm$0.17} \\
& + CutMix + \name & \textbf{5.20}\tiny{$\pm$0.13} & \textbf{5.21}\tiny{$\pm$0.01} & \textbf{5.22}\tiny{$\pm$0.66} & \textbf{4.41}\tiny{$\pm$0.25} & \textbf{4.29}\tiny{$\pm$0.20} & \textbf{5.28}\tiny{$\pm$0.06}\\
\cline{2-8}
& + DDGSD & 5.65\tiny{$\pm$0.14} &5.79\tiny{$\pm$0.08} & 5.96\tiny{$\pm$0.04} & 4.77\tiny{$\pm$0.04} & 4.72\tiny{$\pm$0.13} & 6.83\tiny{$\pm$0.21} \\
& + DDGSD + \name & \textbf{5.31}\tiny{$\pm$0.07} & \textbf{5.52}\tiny{$\pm$0.05} & \textbf{5.75}\tiny{$\pm$0.25} & \textbf{4.45}\tiny{$\pm$0.59} & \textbf{4.14}\tiny{$\pm$0.03} & \textbf{5.74}\tiny{$\pm$0.23} \\
\hline
\multirow{8}{*}{C100}
&
Baseline & 26.37\tiny{$\pm$0.19} & 27.16\tiny{$\pm$0.42} & 28.26\tiny{$\pm$0.18} & 23.64\tiny{$\pm$0.25} & 22.42\tiny{$\pm$0.23} & 28.31\tiny{$\pm$0.35}\\
& + \name &
 \textbf{23.88}\tiny{$\pm$0.06} & \textbf{24.53}\tiny{$\pm$0.13} & \textbf{26.00}\tiny{$\pm$0.03} & \textbf{21.82}\tiny{$\pm$0.19} & \textbf{20.79}\tiny{$\pm$0.33} & \textbf{27.48}\tiny{$\pm$0.21} \\
\cline{2-8}
& + CutOut & 25.76\tiny{$\pm$0.16} & 26.19\tiny{$\pm$0.69} & 27.72\tiny{$\pm$0.02} & 22.09\tiny{$\pm$0.18} & 21.14\tiny{$\pm$0.37} & 28.58\tiny{$\pm$0.47} \\
& + CutOut + \name & \textbf{23.02}\tiny{$\pm$0.08} & \textbf{23.62}\tiny{$\pm$0.17} & \textbf{25.93}\tiny{$\pm$0.28} & \textbf{20.27}\tiny{$\pm$0.33} & \textbf{20.30}\tiny{$\pm$0.01} & \textbf{27.04}\tiny{$\pm$0.43}\\
\cline{2-8}
& + CutMix & 24.07\tiny{$\pm$1.02} & 25.67\tiny{$\pm$0.04} & 27.31\tiny{$\pm$0.42} & 21.42\tiny{$\pm$0.1} & 20.57\tiny{$\pm$0.27} & 
28.27\tiny{$\pm$0.42} \\
&+ CutMix + \name & \textbf{23.48}\tiny{$\pm$0.29} & \textbf{24.06}\tiny{$\pm$0.01} & \textbf{25.77}\tiny{$\pm$0.64} & \textbf{20.93}\tiny{$\pm$0.04} & \textbf{20.10}\tiny{$\pm$0.11} & \textbf{26.82}\tiny{$\pm$0.12}\\
\cline{2-8}
& + DDGSD & 24.36\tiny{$\pm$0.04} & 24.69\tiny{$\pm$0.01} & 26.32\tiny{$\pm$0.23} & 22.55\tiny{$\pm$0.17} & 20.83\tiny{$\pm$0.08} &
27.69\tiny{$\pm$0.32} \\
& + DDGSD + \name & 
\textbf{23.81}\tiny{$\pm$0.11} & \textbf{24.20}\tiny{$\pm$0.28} & \textbf{25.98}\tiny{$\pm$0.04} & \textbf{21.18}\tiny{$\pm$0.42} & \textbf{20.28}\tiny{$\pm$0.01} & \textbf{27.25}\tiny{$\pm$0.28}  \\
\bottomrule
\end{tabular}
}
\end{table*}

%% file: result/r3-table.tex
\begin{table}[!ht]
\caption{
Ablation study on CIFAR-100. 
}\label{table:ablation}
\centering
\resizebox{\linewidth}{!}{
\begin{tabular}{l|ccc} 
\toprule
Networks & Baseline & loss in \eqref{eq:kd-last-batch} removed & All settings\\
\hline
Vgg-16 & 26.37\tiny{$\pm$0.19} & 25.45\tiny{$\pm$0.13} & 23.88\tiny{$\pm$0.06} \\
Vgg-19 & 27.16\tiny{$\pm$0.42} & 25.73\tiny{$\pm$0.28} & 24.53\tiny{$\pm$0.13} \\
ResNet-32 & 28.26\tiny{$\pm$0.18} & 28.01\tiny{$\pm$0.02} & 26.00\tiny{$\pm$0.03} \\
ResNet-110 & 23.64\tiny{$\pm$0.25} & 22.65\tiny{$\pm$0.23} & 21.82\tiny{$\pm$0.19}\\
WRN20-8 & 22.42\tiny{$\pm$0.23} & 21.18\tiny{$\pm$0.01} & 20.79\tiny{$\pm$0.33} \\
DenseNet-40-12 & 28.31\tiny{$\pm$0.35} & 28.30\tiny{$\pm$0.24} & 27.48\tiny{$\pm$0.21} \\
\bottomrule
\end{tabular}
}
\end{table}

%% file: 6-conclusion.tex
\section{Limitation and Conclusion}
In this research, we introduce an efficient self-distillation mechanism for consistency regularization. Without the involvement of a complex pre-trained teacher model or an ensemble of peer students, our method (\name) distills the on-the-fly generated smooth labels in the previous iteration after rearranging the sampling sequence. \name~ regularizes network by imposing training consistency, which is further amplified under data corruption settings. Thus, it boosts the robustness to label noise. Experimental results on three benchmark datasets suggest that our method can consistently outperform the state-of-the-art self distillation mechanism. Moreover, \name, as a universal regularization approach, works in conjunction with augmentations techniques, bringing additional performance gain. However, due to the limitation of computational resources, we did not evaluate the performance on large-scale datasets such as ImageNet, which is left to future work. Additionally, as our method depends on the knowledge transmission between soft labels, in this paper, we focus primarily on classification, which yields another future direction in the extension to other tasks e.g. semantic segmentation, object detection.